\documentclass[a4paper,twoside]{article}
\usepackage{expl3}
\ExplSyntaxOn
\tex_let:D \c_minus_one \scan_stop:
\int_const:Nn \c_minus_one {-1}
\ExplSyntaxOff
\usepackage{epsfig}
\usepackage{subcaption}
\usepackage{calc}
\usepackage{amssymb}
\usepackage{amstext}
\usepackage{amsmath}
\usepackage{outlines}
\usepackage{amsthm}
\usepackage{graphicx}

\usepackage{multicol}
\usepackage{pslatex}
\usepackage{apalike}
\usepackage{color}
\usepackage{xspace}
\usepackage{tabularx}
\usepackage{adjustbox}
\usepackage{acro}
\usepackage[pagebackref=true,breaklinks=true,letterpaper=true,bookmarks=false]{hyperref}
\usepackage{mathtools}

\usepackage{SCITEPRESS}     

\makeatletter

\DeclareRobustCommand\onedot{\futurelet\@let@token\@onedot}
\def\@onedot{\ifx\@let@token.\else.\null\fi\xspace}

\def\etal{\emph{et al}\onedot}
\makeatother

\newcolumntype{Y}{>{\raggedright\arraybackslash}X}
\DeclarePairedDelimiter{\norm}{\lVert}{\rVert}
\newcommand*\rot{\rotatebox{90}}

\newif\ifblindreview
\blindreviewtrue

\DeclareAcronym{ST}{
  short = ST ,
  long  = self-training,
  class = abbrev
}
\DeclareAcronym{UDA}{
  short = UDA ,
  long  = Unsupervised Domain Adaptation,
  class = abbrev
}
\DeclareAcronym{I2I}{
  short = I2I ,
  long  = Image-to-Image Translation,
  class = abbrev
}
\DeclareAcronym{SSL}{
  short = SSL ,
  long  = Semi-Supervised Learning,
  class = abbrev
}
\DeclareAcronym{EMA}{
  short = EMA ,
  long  = Exponential Moving Average,
  class = abbrev
}

\begin{document}
\title{Semantically Consistent Image-to-Image Translation for Unsupervised Domain Adaptation}

\ifblindreview
\author{\authorname{Stephan Brehm\sup{1}, Sebastian Scherer\sup{1} and Rainer Lienhart\sup{1}}
\affiliation{\sup{1}Department of Computer Science, University of Augsburg, Universitätsstr. 6a, Augsburg, Germany}
\email{\{stephan.brehm, sebastian.scherer, rainer.lienhart\}@informatik.uni-augsburg.de}
}
\fi

\abstract{
    \ac{UDA} aims to adapt models trained on a source domain to a new target domain where no labelled data is available. In this work, we investigate the problem of \ac{UDA} from a synthetic computer-generated domain to a similar but real-world domain for learning semantic segmentation. 
    We propose a semantically consistent image-to-image translation method in combination with a consistency regularisation method for \ac{UDA}. 
    We overcome previous limitations on transferring synthetic images to real looking images. We leverage pseudo-labels in order to learn a generative image-to-image translation model that receives additional feedback from semantic labels on both domains. 
    Our method outperforms state-of-the-art methods that combine image-to-image translation and semi-supervised learning on relevant domain adaptation benchmarks, i.e., on GTA5 to Cityscapes and SYNTHIA to Cityscapes. 
    }

\onecolumn \maketitle \normalsize \setcounter{footnote}{0} \vfill

\section{\uppercase{Introduction}}
\label{sec:introduction}
The problem of domain adaptation from a synthetic source domain to the real target domain is mainly motivated by the cheap and almost endless possibilities of automated creation of synthetic data. In contrast, data from the real domain is often hard to acquire. This is especially true for most types of labelled data. A common problem in computer vision, for which acquiring real labelled data is exceptionally hard, is semantic segmentation. Semantic segmentation requires annotations at the pixel level. These annotations commonly need to be created manually in a time-consuming process that also demands rigorous and continuous focus from human annotators. Due to this, a reasonable approach is to learn as much as possible from synthetic data and then transfer the knowledge to the real domain. However, convolutional neural networks (CNNs), in general, learn features from the domain on which they are trained on. This causes networks to perform poorly on unseen domains due to the visual gap between these domains. Unsupervised domain adaptation~(UDA) aims to bridge this domain gap in order to learn models that perform well in the target domain without ever using labels from that domain. 

In this work, we aim to improve the quality and usefulness of synthetic data by transforming synthetic images such that they look more like real images. However, such an \ac{I2I} approach cannot change fundamental differences in image content because it needs to keep the transformed images consistent to the semantic labels of synthetic images. Because of this, we cannot bridge the gap between synthetic and real domains solely by an \ac{I2I} approach.

The remaining differences are in image content and include object shapes, object frequency, and differences in viewpoint.
Following recent work, we utilise a \ac{SSL} framework which allows us to include unlabelled data from the real target domain into the training of a semantic segmentation network. 
The main contributions of our work are summarised as follows:
\begin{enumerate}
    \item We propose a semantically consistent \ac{I2I} method that combines an adversarial approach with a segmentation objective that is optimised jointly by both generator and discriminator. 
    \item We improve both adversarial and segmentation objectives with self-supervised techniques that include the unlabelled data from the real target domain into the training. Our method outperforms state-of-the-art combinations of \ac{I2I} and \ac{SSL} on challenging benchmarks for \ac{UDA}. We provide extensive experiments and analyses and show which components are essential to our approach.
\end{enumerate}

\section{Related Work}
\paragraph{Semantic Segmentation} is the task of labelling every pixel of an image according to the object class it belongs to. In 2015, Long~\etal proposed fully convolution neural networks~(FCNs)~\cite{long2015fully} improving massively upon classical methods for semantic segmentation. Since then, many new methods based on deep convolutional neural networks were proposed~\cite{chen2017deeplab,wang2020deepHR,yu2016multi}. However, these methods are directly learned on the real target domain in a supervised fashion. In contrast, our method is able to learn about the real domain without the necessity of annotated real data.

\paragraph{Image-to-Image translation (I2I)} methods~\cite{choi2019self,pizzati2020domain} have been widely used to bridge the gap between synthetic and real data. For image data, this is commonly achieved by learning a deep convolutional neural network that receives a synthetic image from the source domain as input and manipulates it in such a way that it looks more realistic. Basically, these methods try to make the synthetic data look more real. For our task, it is important that the manipulated versions of the synthetic data can be used for supervised training of a segmentation method subsequently. In order to learn useful manipulations for such a task, these systems need mechanisms that keep the overall image consistent with the synthetic labels.

\paragraph{Unsupervised Domain Adaptation (UDA)} for semantic segmentation has been extensively studied in the last years. 
Adversarial training is often used in \ac{UDA} methods to adapt either input space, feature space or output space of a semantic segmentation network~\cite{toldo2020unsupervised_survey}. On the feature level, a discriminator is trained to distinguish between feature maps from different domains~\cite{hoffman2016fcns,chen2017no,hong2018conditional}. Popular input space adaptation techniques utilise frameworks based on cycle consistency~\cite{CycleGAN} for \ac{UDA}~\cite{CyCada,sankaranarayanan2018learning,chen2019crdoco,murez2018image}. However, cycle consistency allows almost arbitrary transformations as long as they can be reversed. In contrast to these methods, we perform input space adaptation in our \ac{I2I} method without the need for cycle consistency. Our method simply uses the annotations of the synthetic data to directly enforce consistency.

\paragraph{Semi-supervised learning (SSL)} aims to include unlabelled data into the training which allows to use data from the real domain. The dominant approaches for \ac{SSL} are pseudo-labelling and consistency regularisation. An extensive overview can be found in this survey~\cite{SSL_Survey}.
Pseudo-Labeling was first proposed in~\cite{Pseudo-Labels}. Xie~\etal~\cite{Pseudo-Labels_Teacher-Student} recently showed that pseudo-labels can indeed improve overall performance on image classification tasks. They train a network on labelled data and reuse high-confidence predictions on unlabelled data as pseudo-labels. Pseudo-labels are then included in the full dataset on which a new model is subsequently trained from scratch.

Many recent \ac{SSL} methods~\cite{ke2019dual,MeanTeacher} include consistency regularisation. They employ unlabelled data to produce consistent predictions under different perturbations. Possible perturbations can be data augmentation, dropout or simple noise on the input data. The trained model should be robust against such perturbations. These approaches leverage the idea that a classifier should output the same distribution for different augmented versions of an unlabelled sample. This is typically achieved by minimising the difference between the prediction of a model on different perturbed versions of the same input. We utilise such an approach to boost our performance even further. Our \ac{I2I} translation module also leverages pseudo-labels created by such a consistency regularised method.

\section{Method}
In this work, we assume that image data is available from both the synthetic \textit{source} domain as well as the real \textit{target} domain. However, annotations are only available in the synthetic \textit{source} domain. We aim to bridge the gap between the \textit{source} and the \textit{target} data. Our image-sets are sampled from their respective domains. The first step is to bring the two domains closer together visually which means transforming the synthetic data such that it looks more real. By keeping consistency with the labels of the synthetic source data, we are able to generate a dataset of images with corresponding labels that closely resembles the real target data in terms of colour, texture and lighting.
Basically, we aim to improve domain adaptation by transferring the style of the real target domain onto images of the synthetic source domain. However, fundamental differences in image composition cannot be learned with such an \ac{I2I} method. For this reason, we use the translated synthetic data in conjunction with unlabelled data from the real \textit{target} domain to train our segmentation model. 
We propose a training process that consists of three phases.
\begin{itemize}
    \item (a) In the \textit{warm-up phase} the \ac{SSL} method is used to train an initial segmentation model $M_0$ as the initial pseudo label generator.
    \item (b) In the \textit{\ac{I2I} training phase}, we use the pseudo labels from model $M_0$ and train our generator $G$ to produce real looking images from synthetic images.
    \item (c) In the \textit{segmentation training phase}, we combine the \ac{SSL} method with the translated images by generator $G$ and train our new segmentation network $M_1$.
\end{itemize}

\subsection{Notation}
In the following, we will denote $\mathcal{S}$ as the \textit{source} domain and $\mathcal{T}$ as the \textit{target} domain. $X_\mathcal{S}$ and $X_\mathcal{T}$ are sets of images sampled from $\mathcal{S}$ and $\mathcal{T}$, i.e., synthetic images and real images, respectively. $\mathbf{x}^s \in X_\mathcal{S}$ and $\mathbf{x}^t \in X_\mathcal{T}$ are images sampled from their respective image sets $X_\mathcal{S}$ and $X_\mathcal{T}$. 

Both domains $\mathcal{S}$ and $\mathcal{T}$ share a common set of categories $C$.
We denote the label of an image $\mathbf{x}^{d}$ of domain $d\in \{\mathcal{S},\mathcal{T}\}$ as $\mathbf{y}^{d}$. In this work $\mathbf{y}^{s}\in Y_\mathcal{S}$ generally is a synthetic segmentation mask and $\mathbf{\hat{y}}^{t}\in \hat{Y}_\mathcal{T}$ is a segmentation mask that contains pseudo-labels. For the purpose of this work, the real segmentation labels $\mathbf{y}^{t}\in Y_\mathcal{T}$ are not available for training.
$\mathcal{L}^{W,n}_{i,j,c}(\mathbf{x},\mathbf{y})$ denotes a loss function based on image $\mathbf{x}$ and label $\mathbf{y}$ that is used to update trainable parameters of a network $W$. $n$ is an identifier/name for the loss function. Note that we often omit the arguments $(\mathbf{x},\mathbf{y})$ in order to simplify our notation. Also note that $W$ only indicates which network is updated by $\mathcal{L}^{n}$ and thus may also be omitted in more general statements and definitions. $i,j,c$ are indices that we use to refer to specific dimensions and or individual values in the calculated loss if necessary. In general, we assume multidimensional arrays to be in the order of $height\times width\times channels$. 

In this work, we aim to train a segmentation network $F_\mathcal{S}$ on $X_\mathcal{S}$~(and possibly $X_\mathcal{T}$) that estimates $Y_\mathcal{T}$ without the need for any $\mathbf{y}^{t}\in Y_\mathcal{T}$ during training, i.e., we aim to generalise from domain $\mathcal{S}$ to domain $\mathcal{T}$ without ever using any labels from $\mathcal{T}$.

\subsection{Image-to-Image Translation}
\label{sec:translation}
\begin{figure*}
    \includegraphics[width=\linewidth]{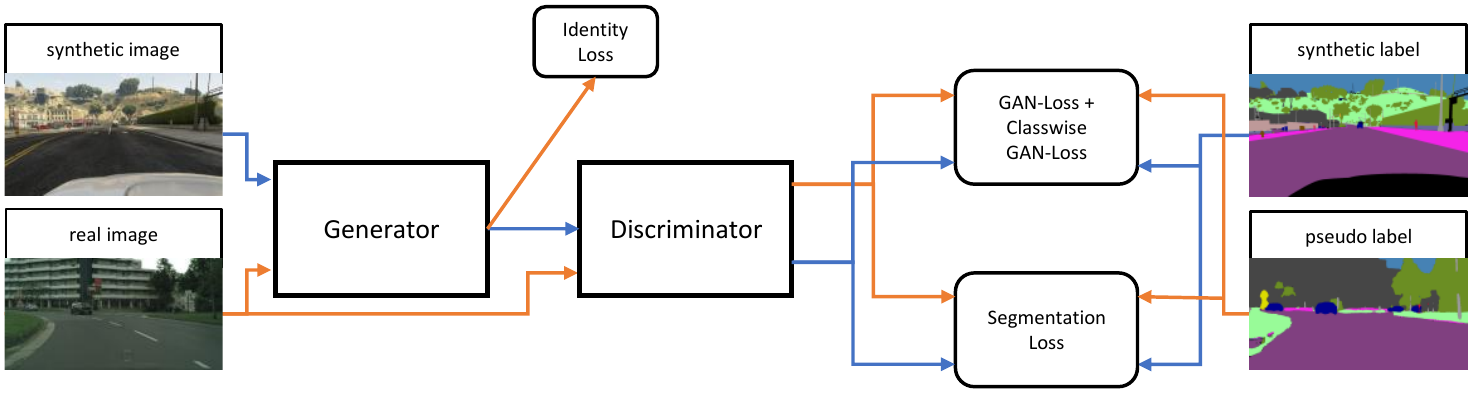}
    \caption{Schematic illustration of our proposed \ac{I2I} method. The colored arrows visualize data-flow.}
    \label{fig:translationmethod}
\end{figure*}
In order to generalise to a target domain $\mathcal{T}$ we need to bring our source images $X_\mathcal{S}$ closer to $\mathcal{T}$.
In this section, we detail our approach to this goal. \autoref{fig:translationmethod} illustrates the employed architecture.
We aim to transfer images from a synthetic domain to the real domain, because the synthetic domain allows to easily create images and corresponding labels by simulating an environment that is similar to the real environment. The transformed images are supposed to deliver maximum performance on real data when learning a fully convolutional method for semantic segmentation. In order to achieve this, such a transformation needs to meet two major requirements.
\begin{enumerate}
    \item Transformed images need to look as realistic as possible
    \item Transformed images need to maintain consistency to the segmentation labels
\end{enumerate}
We propose an \ac{I2I} method that consists of two networks in an adversarial setting. Our learning task is designed to enforce the requirements given above. Requirement 1. is tackled by an adversarial objective. For Requirement 2., we extend the adversarial setting with an additional cooperative segmentation objective that is jointly optimised by both adversaries. Following, we give a brief overview of the employed generator and discriminator architectures.

\paragraph{Generator}
\label{sec:generator}
We use a simple encoder-decoder architecture with strided convolutions for down-sampling the input image by a factor of 8. These down-sampling operations are followed by a single residual block. The decoder is simply a stack of deconvolution layers. Note that we do not include any skip/residual connections between encoder and decoder.
In every layer, we append a learned scaling factor as well as a learned bias term. We use leaky-ReLU activations for all layers except the output layer. We also use deconvolution kernels, PixelNorm, Equalized Learning Rate, Adaptive Instance Normalization and Stochastic Variation as proposed in~\cite{karras2017progressive}.
Similar to Taigman~\etal~\cite{taigman2016unsupervised}, we incorporate images from the target domain in the training of the generator by adding an identity objective. Thus, the generator learns to encode and reconstruct real images in addition to its main \ac{I2I} task. The additional feedback is a way to directly learn about the structure and texture of real images.

\paragraph{Discriminator}
\label{sec:discriminator}
Instead of discriminating between generated images and real images directly, we propose to use prior knowledge for discrimination. We use image-features of an ImageNet-pretrained~\cite{deng2009imagenet} VGG16~network~\cite{simonyan2014vgg} after the \textit{block3\_conv3} layer. We build additional 3 residual blocks~\cite{he2016deep} on top of these features. Note that we do not fine-tune the pre-trained VGG16 part of the discriminator.
Our discriminator features two distinct outputs which are each computed on top of a separate decoder, as illustrated in \autoref{fig:translationmethod}. We refer to these decoders as the GAN-head~\cite{goodfellow2014gan} and the auxiliary classifier head~(AC-head)~\cite{odena2017conditional}. Both heads output results at image resolution.
The task of the AC-head is to learn semantic segmentation in the target domain. Thus we feed the translated images in conjunction with the labels of the synthetic data as well as real images with corresponding pseudo-labels. Note that these pseudo-labels can be created online during training. 
However, using pseudo-labels created by our proposed segmentation system that is detailed in section~\ref{sec:segmentation} generally improved results.
The AC-head is important in many ways. First, the segmentation feedback generated in the AC-head forces previous layers to learn segmentation-specific features which in turn helps the GAN-head to discriminate between images that are translated from the synthetic domain and images from the real domain.
Second, it provides consistency feedback for the generator. We achieve this by propagating gradients based on the segmentation loss of the AC-head into the weights of the discriminator as well as into the weights of the generator. This effectively optimises a joint segmentation objective in both generator and discriminator. Thus, the generator is punished if the translated images are not consistent with the label of the synthetic input data. 
Third, it can be used to generate pseudo-labels.
The GAN-Head is responsible for the discrimination between real examples and fake examples. Note that GAN-Heads are commonly learned with a single label only~(\textit{real} or \textit{fake}). We extend this formulation to include class information. Thus, allowing the discriminator to directly learn to compare images on a class specific level. We utilize segmentation labels in the GAN-head to create this class-specific feedback. In GAN-terminology this means that we treat pseudo-labels as real and the synthetic labels as fake. For each class, we produce an additional output feature map. On this feature map, feedback is only applied at positions that belong to the given object class as defined by either synthetic labels or pseudo-labels. 
 
\paragraph{Optimisation}
\label{sec:im2imoptimization}
Given the supervised segmentation loss $\mathcal{L}^{seg}(D(G(\mathbf{x}^s)), \mathbf{y}^{s})$~(softmax cross-entropy) based on synthetic labels and the supervised segmentation loss $\mathcal{L}^{seg}(D(\mathbf{x}^t),\mathbf{\hat{y}}^{t})$ based on the pseudo-labels, we compute the total segmentation loss $\mathcal{L}^{seg}$ as given in \autoref{eg:segmloss}. We use the symmetric cross-entropy loss~\cite{wang2019symmetric} for $\mathcal{L}^{seg}(D(\mathbf{x}^t),\mathbf{\hat{y}}^{t})$ because it increases robustness when using noisy pseudo-labels $\mathbf{\hat{y}}^{t}$.
\begin{equation}\label{eg:segmloss}
    \mathcal{L}^{seg} = \mathcal{L}^{seg}(D(G(\mathbf{x}^s)), \mathbf{y}^{s}) + \lambda^{pl}\mathcal{L}^{seg}(D(\mathbf{x}^t), \mathbf{\hat{y}}^{t})
\end{equation}
where $\lambda^{pl}$ is a scaling factor that we use to control the relative impact of the pseudo-labels on the overall error of the model. 

Like the segmentation loss, we calculate the GAN-Loss $\mathcal{L}^{gan}$ on a pixel-wise basis, i.e., the GAN-loss is the average of all losses computed over all pixels $I_{i,j}$ of an input image $I$. 
Let $m(a,b)$ be a derivable distance metric between $a$ and $b$. We give a general error function $\mathcal{L}^{D,dgan}$ for a discriminator $D$ as well as a general error function $\mathcal{L}^{G,dgan}$ for a generator $G$ in \autoref{eg:ganloss1} and \autoref{eg:ganloss2} respectively.
\begin{align}
    \label{eg:ganloss1}
    \mathcal{L}^{D,dgan}_{i,j} &= m(D(G(\mathbf{x}^s))_{i,j}, 0) + m(D(\mathbf{x}^t)_{i,j}, 1)\\
    \label{eg:ganloss2}
    \mathcal{L}^{G,dgan}_{i,j} &= m(D(G(\mathbf{x}^s))_{i,j}, 1)
\end{align}
We use the mean-squared error for the distance metric $m$.

Now let, $\mathbf{y}^{s}$ be the one-hot encoded segmentation label for a given image $\mathbf{x}^{s}$, i.e., $\mathbf{y}^{s}_{i,j,c}$ equals one if $I_{i,j}$ belongs to class $c$. Otherwise $\mathbf{y}^{s}_{i,j,c}$ is zero. The same applies to the pseudo-label $\mathbf{\hat{y}}^{t}$. We define a class-wise and pixel-wise error $\mathcal{L}^{D,cgan}_{i,j,c}$ for discriminator $D$ in \autoref{eg:clsganloss1} and a class-wise and pixel-wise error $\mathcal{L}^{G,cgan}_{i,j,c}$ for generator $G$ in \autoref{eg:clsganloss2}.
\begin{align}
    \label{eg:clsganloss1}
    \mathcal{L}^{D,cgan}_{i,j,c} &= m(D(G(\mathbf{x}^s))_{i,j,c}, 0) \cdot \mathbf{y}^{s}_{i,j,c} +\\
    \nonumber&+ m(D(\mathbf{x}^t)_{i,j,c}, 1) \cdot \mathbf{\hat{y}}^{t}_{i,j,c}\\
    \label{eg:clsganloss2}
    \mathcal{L}^{G,cgan}_{i,j,c} &= m(D(G(\mathbf{x}^s))_{i,j,c}, 1) \cdot \mathbf{y}^{s}_{i,j,c}
\end{align}
Hence, the full adversarial loss for $D$ is:

\begin{align}\label{eg:gantotalloss}
    \nonumber&\mathcal{L}^{D,gan} = \\
    &\sum_{i=0}^H\sum_{j=0}^W \left(\mathcal{L}^{D,dgan}_{i,j} + \frac{\lambda^{cgan}}{|C|}\sum_{c \in C} \mathcal{L}^{D,cgan}_{i,j,c}\right).
\end{align}
where $\lambda^{cgan}$ is a scaling factor that we use to control the relative impact of the class-wise loss on the overall error of the model,
$H$ is the height and $W$ is the width of both images and labels. $\mathcal{L}^{G,gan}$ can be computed analogous.

The total loss of the discriminator $\mathcal{L}^{D}$ is a combination of segmentation and GAN losses as given in \autoref{eg:discloss}.
\begin{equation}
    \label{eg:discloss}
    \mathcal{L}^{D} = \frac{1}{HW}\left(\mathcal{L}^{D,seg} + \mathcal{L}^{D,gan}\right) \\
\end{equation}
With $\mathcal{L}^{G, id}$ being the identity reconstruction error on a real image $\mathbf{x}^t$ from the target domain, we compute the generator loss $\mathcal{L}^{G}$ as given in \autoref{eq:genloss}.
\begin{align}
    \mathcal{L}^{G, id} = & \sum_{i=0}^{H}\sum_{j=0}^{W} \norm{G(\mathbf{x}^t)_{i,j} - \mathbf{x}^t_{i,j}}_1 \\
    \label{eq:genloss}
    \mathcal{L}^{G} = &\frac{1}{HW}\left(\mathcal{L}^{G,seg} + \mathcal{L}^{G,gan} + \mathcal{L}^{G, id}\right)
\end{align}

Note that the second term of $\mathcal{L}^{seg}$ from \autoref{eg:segmloss} is not affected by $G$ and thus $\mathcal{L}^{G,seg}$ is reduced to the first term of $\mathcal{L}^{seg}$ when back-propagating. This means that $G$ should generate images that match the synthetic segmentation labels. On the other hand, $\mathcal{L}^{G,gan}$ incentivises more realistic images. By using $\mathcal{L}^{G, id}$ we are able to show real images from the target domain to $G$. This combination of various objectives enables us to learn an \ac{I2I} model that satisfies the requirement of realistic transformation while simultaneously keeping consistency with the synthetic labels. We use the resulting images to subsequently learn a better segmentation model.

\subsection{Segmentation Training}
\label{sec:segmentation}
At this point, we are able to transform synthetic images into real images while keeping the semantic content of the image unchanged. Thus, we can use the transformed images with the original synthetic annotations. However, some issues remain to be solved. An \ac{I2I} method cannot bridge certain differences between the domains. The remaining differences are in image content and include object shapes, object frequency, and differences in viewpoint which can not be learned by our \ac{I2I} module. As a solution, we incorporate images from the real domain directly into the training of the segmentation model via a SSL framework. 

Similar to Tarvainen~\etal~\cite{MeanTeacher}, we make use of two networks: a student network $F_S$ and a teacher network $F_T$. The architecture of the teacher network is identical to the one of the student network. The weights of the teacher model are an \ac{EMA} of the student's weights. 
Given an image from the target domain, the teacher's prediction serves as a label for the student, forcing consistency in the prediction of both models under different perturbations.

The overall objective $\mathcal{L}^{F_S}$ is a combination of a supervised segmentation loss $\mathcal{L}^{F_S,seg}$ and the self-supervised consistency loss $\mathcal{L}^{F_S, con}$ as detailed in \autoref{eq:overall}.
\begin{equation}
    \mathcal{L}^{F_S} = \mathcal{L}^{F_S,seg}(\mathbf{x}^s, \mathbf{y}^s) + \lambda^{con} \mathcal{L}^{F_S, con}(\mathbf{x}^t),
    \label{eq:overall}
\end{equation}
where $\lambda^{con}$ is a trade-off parameter. 

For the supervised training, we incorporate synthetic images $\mathbf{x}^s \in X_\mathcal{S}$ and a transformed version obtained from the generator $G(\mathbf{x}^s)$ of our proposed \ac{I2I} method. We calculate a combined loss $\mathcal{L}^{F_S, seg}$ as given in \autoref{eq:Sup}
\begin{equation}
    \mathcal{L}^{F_S, seg} =  
    \mathcal{L}^{seg}((F_S(\mathbf{x}^s)), \mathbf{y}^{s}) 
    +  \mathcal{L}^{seg}((F_S(G(\mathbf{x}^s))), \mathbf{y}^{s}),
    \label{eq:Sup}
\end{equation}
where $\mathcal{L}^{seg}$ again is the softmax cross-entropy error.

For the semi-supervised training, we incorporate images from the real domain $\mathbf{x}^t \in X_\mathcal{T}$. We calculate the consistency loss $\mathcal{L}^{F_S, con}(\mathbf{x}^t)$ as given in \autoref{eq:SSL}.
\begin{equation}
        \mathcal{L}^{F_S, con}=  ||\sigma(F_T(\mathbf{x}^t)) - \sigma(F_S(\mathbf{P}(\mathbf{x}^t))) ||^2_2, 
    \label{eq:SSL}
\end{equation}
where $\sigma$ is the softmax activation function and $\mathbf{P}(\mathbf{x}^t)$ is a strongly perturbed version of the input image $\mathbf{x}^t$. Similar to~\cite{zhou2020uncertainty}, we utilise color jittering, Gaussian blurring and noise as perturbations on $\mathbf{x}^t$. 

As our \ac{I2I} model can also leverage pseudo-labels, we perform an identical training with a supervised loss and a consistency loss, but without transformed images $G(\mathbf{x}^s)$ in the \textit{warm-up phase}. We use the resulting model to generate pseudo-labels that we use to train $G$ in the \textit{\ac{I2I} training phase}. We then use $G$ to translate synthetic images. These translated images in conjunction with the original synthetic images constitute the training data for the \textit{segmentation training phase}.

\begin{table*}
\caption{Results of domain adaptation from GTA5 to Cityscapes using a VGG backbone.}
    \resizebox{\linewidth}{!}{
    \begin{tabularx}{21cm}{l|| *{19}{Y} | *{1}{Y}} 
     &  \rot{road} &  \rot{sidewalk} &  \rot{building} &  \rot{wall} &  \rot{fence} &  \rot{pole} &  \rot{traffic light}  &  \rot{traffic sign} &  \rot{vegetation} &  \rot{terrain} &  \rot{sky} &  \rot{person} &  \rot{rider} &  \rot{car} &  \rot{truck} &  \rot{bus} &  \rot{train} &  \rot{motorbike} &  \rot{bicycle} &  \rot{$\textbf{mIoU}^{19}$} \\ \hline

         Source only (ours) &  79.2 &  30.6 &  76.0 &  22.7 &  11.1 &  19.2 &  11.0 &  2.2 &  80.3 &  30.4 &  73.5 &  39.3 &  0.5 &  75.3 &  17.3 &  9.5 &  0.0 &  1.4 &  0.0 &  30.5  \\\hline\hline
     
         CyCADA \cite{CyCada} &  85.2 &  37.2 &  76.5 &  21.8 &  15.0 &  23.8 &  22.9 &  21.5 &  80.5 &  31.3 &  60.7 &  50.5 &  9.0 &  76.9 &  17.1 &  28.2 &  \textbf{4.5} &  9.8 &  0.0 &  35.4 \\ 
      
         CBST-SP \cite{cbst} &  90.4 &  50.8 &  72.0 &  18.3 &  9.5 &  27.2 &  28.6 &  14.1 &  82.4 &  25.1 &  70.8 &  42.6 &  14.5 &  76.9 &  5.9 &  12.5 &  1.2 &  14.0 &  28.6 &  36.1   \\

         DCAN \cite{wu2018dcan} &  82.3 &  26.7 &  77.4 &  23.7 &  20.5 &  20.4 &  30.3 &  15.9 &  80.9 &  25.4 &  69.5 &  52.6 &  11.1 &  79.6 &  24.9 &  21.2 &  1.3 &  17.0 &  6.7 &   36.2   \\ 

         SWD \cite{lee2019SWD} &  91.0 &  35.7 &  78.0 &  21.6 &  21.7 &  31.8 &  30.2 &  25.2 &  80.2 &  23.9 &  74.1 &  53.1 &  15.8 &  79.3 &  22.1 &  26.5 &  1.5 &  17.2 &  \textbf{30.4} &  39.9   \\ 
        
         SIM \cite{Wang_2020_CVPR_Sim} &  88.1 &  35.8 &  83.1 &  25.8 &  23.9 &  29.2 &  28.8 &  28.6 &  83.0 &  36.7 &  82.3 &  53.7 &  22.8 &  82.3 &  26.4 &  38.6 &  0.0 &  19.6 &  17.1 &  42.4   \\ 
        
         TGCF-DA + SE \cite{choi2019self} &  90.2 &  51.5 &  81.1 &  15.0 &  10.7 &  \textbf{37.5} &  35.2 &  28.9 &  84.1 &  32.7 &  75.9 &  62.7 &  19.9 &  82.6 &   22.9 &  28.3 &  0.0 &  23.0 &  25.4 &  42.5 \\ 
    
         FADA \cite{wang2020classes} &  92.3 &  51.1 &  83.7 &  33.1 &  \textbf{29.1} &  28.5 &  28.0 &  21.0 &  82.6 &  32.6 &  85.3 &  55.2 &  28.8 &  83.5 &  24.4 &  37.4 &  0.0 &  21.1 &  15.2 &  43.8   \\ 
        
         PCEDA \cite{yang2020phase} &  90.2 &  44.7 &  82.0 &  28.4 &  28.4 &  24.4 &  33.7 &  35.6 &  83.7 &  40.5 &  75.1 &  54.4 &  28.2 &  80.3 &  23.8 &  39.4 &  0.0 &  22.8 &  30.8 &  44.6   \\ 
    
         Zhou \etal \cite{zhou2020uncertainty} &  \textbf{95.1} &  \textbf{66.5} &  84.7 &  35.1 &  19.8 &  31.2 &  35.0 &  32.1 &  86.2 &  \textbf{43.4} &  82.5 &  61.0 &  25.1 &  87.1 &  \textbf{35.3} &  \textbf{46.1} &  0.0 &  24.6 &  17.5 &  47.8  \\ \hline
      
         Ours &  94.4 &  65.3 &  \textbf{85.9} &  \textbf{39.0} &  22.2 &  35.4 &  \textbf{39.1} &  \textbf{37.3} &  \textbf{86.7} &  42.3 &  \textbf{88.1} &  \textbf{62.7} &  \textbf{36.2} &  \textbf{87.6} &  33.8 &  45.0 &  0.0 &  \textbf{26.5} &  24.2 &  \textbf{50.1} \\  \hline\hline
      
         Target only (ours) &  96.9 &  79.7 &  89.6 &  46.0 &  47.6 &  47.4 &  55.9 &  64.5 &  90.0 &  60.7 &  91.7 &  72.4 &  49.1 &  92.4 &  56.7 &  75.4 &  54.0 &  51.0 &  69.4 &  68.0 \\ \hline

    \end{tabularx}
    }
\label{tab:gta5}
\end{table*}

\begin{table*}
\caption{Results of domain adaptation from SYNTHIA to Cityscapes using a VGG backbone. $\textbf{mIoU}^{16}$ and $\textbf{mIoU}^{13}$ are computed on 16 and 13 classes respectively. * means that classes are included in $\textbf{mIoU}^{16}$ but excluded in $\textbf{mIoU}^{13}$.}
    \resizebox{\linewidth}{!}{
    \begin{tabularx}{21cm}{l|| *{16}{Y} | *{1}{Y} | *{1}{Y}}
     &  \rot{road} &  \rot{sidewalk} &  \rot{building} &  \rot{wall*} &  \rot{fence*} &  \rot{pole*} &  \rot{traffic light}  &  \rot{traffic sign} &  \rot{vegetation} &  \rot{sky} &  \rot{person} &  \rot{rider} &  \rot{car} &  \rot{bus} &  \rot{motorbike} &  \rot{bicycle} &  \rot{$\textbf{mIoU}^{16}$} &  \rot{$\textbf{mIoU}^{13}$} \\ \hline
         Source only (ours) &  42.0 &  19.6 &  60.4 &  6.3 &  0.1 &  28.3 &  2.1 &  10.3 &  76.2 &  76.0 &  44.3 &  7.2 &  62.5 &  14.8 &  3.2 &  10.6 &  29.0  &  33.0 \\ \hline \hline
        
         DCAN \cite{wu2018dcan} &  79.9 &  30.4 &  70.8 &  1.6 &  \textbf{0.6} &  22.3 &  6.7 &  23.0 &  76.9 &  73.9 &  41.9 &  16.7 &  61.7 &  11.5 &  10.3 &  38.6 &  35.4 & - \\ 
      
          CBST \cite{cbst} &  69.6 &  28.7 &  69.5 &  \textbf{12.1} &  0.1 &  25.4 &  11.9 &  13.6 &  82.0 &  81.9 &  49.1 &  14.5 &  66.0 &  6.6 &  3.7 &  32.4 &  35.4 &  40.7\\ 
      
         SWD \cite{lee2019SWD} &  83.3 &  35.4 &  \textbf{82.1} &  - &  - &  - &  \textbf{12.2} &  12.6 &  \textbf{83.8} &  76.5 &  47.4 &  12.0 &  71.5 &  17.9 &  1.6 &  29.7 &  - &  43.5 \\ 
      
         FADA \cite{wang2020classes} &  80.4 &  35.9 &  80.9 &  2.5 &  0.3 &  \textbf{30.4} &  7.9 &  22.3 &  81.8 &  \textbf{83.6} &  48.9 &  16.8 &  77.7 &  \textbf{31.1} &  13.5 &  17.9 &  39.5 &  46.0 \\ 

         TGCF-DA + SE \cite{choi2019self} &  90.1 &  48.6 &  80.7 &  2.2 &  0.2 &  27.2 &  3.2 &  14.3 &  82.1 &  78.4 &  \textbf{54.4} &  16.4 &  82.5 &  12.3 &  1.7 &  21.8 &  38.5 &  46.6 \\
        
         PCEDA \cite{yang2020phase} &  79.7 &  35.2 &  78.7 &  1.4 &  \textbf{0.6} &  23.1 &  10.0 &  \textbf{28.9} &  79.6 &  81.2 &  51.2 &  \textbf{25.1} &  72.2 &  24.1 &  \textbf{16.7} &  \textbf{50.4} &  41.1 &  48.7 \\
        
         Zhou \etal \cite{zhou2020uncertainty} &  93.1 &  53.2 &  81.1 &  2.6 &  \textbf{0.6} &  29.1 &  7.8 &  15.7 &  81.7 &  81.6 &  53.6 &  20.1 &  82.7 &  22.9 &  7.7 &  31.3 &  41.5 &  48.6   \\ \hline

         Ours &  \textbf{94.8} &  \textbf{67.2} &  81.9 &  6.1 &  0.1 &  29.6 &  0.1 &  19.7 &  82.2 &  81.1 &  50.2 &  17.0 &  \textbf{84.6} &  30.8 &  12.4 &  25.1 &  \textbf{42.7} &  \textbf{49.8} \\ \hline\hline
 
         Target only (ours)  &  96.9 &  79.7 &  89.6 &  46.0 &  47.6 &  47.4 &  55.9 &  64.5 &  90.0 &  91.7 &  72.4 &  49.1 &  92.4 &  75.4 &  51.0 &  69.4 &  70.0 &  75.4 \\ \hline
    
    \end{tabularx}
    }
\label{tab:synthia}
\end{table*}

\section{Experiments and Results}
In this section, we detail experiments that we conducted in order to show the performance of our proposed methods. In Section \ref{sec:datasets} we shortly introduce the datasets that we used. In Section \ref{sec:implementationdetails} we give important details on the training and validation protocols. In Section \ref{sec:Comp_SotA} we compare our methods to previous state-of-the art methods on two public benchmarks. In Section \ref{sec:ablation} we conduct an ablation study to show the impact of individual components on our results.
\subsection{Datasets}
\label{sec:datasets}
Following common practice for unsupervised domain adaptation in semantic segmentation, we use the GTA5~\cite{Richter2016gta5} and SYNTHIA~\cite{ros2016synthia} datasets as our synthetic domain and the Cityscapes dataset~\cite{Cordts2016Cityscapes} as our real domain. The datasets are detailed below.

\paragraph{Cityscapes} dataset~\cite{Cordts2016Cityscapes} contains images of urban street scenes collected around Germany and neighbouring countries. It consists of a training set with 2975 images and a validation set of 500 images. We report $\textbf{mIoU}^{19}$ for the 19 object classes that are annotated. We also use the available 89250 unlabeled images for learning our \ac{I2I} method. 
\paragraph{GTA5} dataset~\cite{Richter2016gta5} contains images rendered by the game Grand Theft Auto 5. It consists of 24966 images with corresponding pixel-level semantic segmentation annotations and a set of object classes that is compatible to the annotations of the Cityscapes dataset. 
\paragraph{SYNTHIA} dataset~\cite{ros2016synthia} consists of a collection of images rendered from a virtual city. We use the SYNTHIA-RAND-CITYSCAPES subset, which consists of 9400 images with pixel-wise annotations. Note that the \textit{terrain}, \textit{truck}, \textit{train} classes are not annotated in the SYNTHIA dataset. Thus, we use the remaining 16 classes that are common with the Cityscapes dataset. 
We evaluate $\textbf{mIoU}^{16}$ on these 16 classes and $\textbf{mIoU}^{13}$ on a subset of 13 classes. $\textbf{mIoU}^{13}$ is a common metric for evaluation on the SYNTHIA dataset that excludes certain classes that are especially hard and/or underrepresented in the data.
\subsection{Implementation details} 
\label{sec:implementationdetails}

For a fair comparison to previous work~\cite{choi2019self,zhou2020uncertainty}, we adopt the VGG16 backbone~\cite{simonyan2014vgg} pre-trained on the ImageNet dataset~\cite{deng2009imagenet}. Following Deeplab-v2~\cite{chen2017deeplab}, we incorporate Atrous Spatial Pyramid Pooling~(ASPP) as the decoder and use an bi-linear up-sampling to get the segmentation output at image resolution. We use this model for all our experiments. We use the Adam optimiser~\cite{kingma2015adam} for the segmentation training with an initial learning rate of $1 \times 10^{-5}$ and exponential weight decay. The generator and discriminator are trained with a constant learning rate of $1 \times 10^{-5}$ for 1 million iterations. For validation purposes, we keep exponential moving averages of the generator weights. We use a 50/50 split of synthetic and transformed data to learn the segmentation model. All networks are trained with gradient clipping at a global norm of $5$. We set all EMA decay values to 0.999. During the first 10.000 training steps of $F_S$, we keep $\lambda^{con}$ to zero. We report results of the teacher network which is a smoothed version of the trained segmentation network. We fade $\lambda^{pseudo}$ and $\lambda^{cgan}$ linearly from zero at iteration 20,000 to 0.3 at iteration 100,000 when learning with pseudo-labels created online. When learning with pre-computed pseudo-labels from our proposed segmentation model, we keep $\lambda_{pseudo}$ and $\lambda_{cgan}$ at 0.3 at all times. All images in all experiments are re-scaled such that the longer side is 1024 pixels. For training, we crop $384\times 384$ regions from these images.

\subsection{Comparisons with the State-of-the-Art}
\label{sec:Comp_SotA}
We compare the results of our method to state-of-the-art methods that combine \ac{I2I} and \ac{SSL} on the two standard benchmarks: “GTA5~to~Cityscapes” and “SYNTHIA~to~Cityscapes” in \autoref{tab:gta5} and \autoref{tab:synthia}, respectively. Our method improves upon current state-of-the art in both benchmarks. Like many other methods, we suffer from fundamental differences in the definition of certain objects, e.g., the \textit{train}~class in the Cityscapes dataset which includes objects like street-cars is actually very similar to the \textit{bus} class in the GTA5~dataset~\cite{zhang2017curriculum}. Such a difference obviously impacts performance on both classes. Indeed, due to the similarity of buses in the source data to trains in the target data, we would actually expect the converted buses to incorporate certain features of the target trains. During segmentation training, we then actively enforce predictions for the bus class which ultimately results in poor performance. We identified an additional issue that affects the \textit{car} and \textit{truck} annotations in the GTA5~dataset. Here, one of the synthetic car models is consistently annotated with the \textit{truck} label. We think that our method that directly enforces label consistency is overly prone to such differences and errors in the annotations. Also note that domain adaptation from SYNTHIA to Cityscapes, in general, is inferior to the variant trained on the GTA5 dataset. The domain gap between the SYNTHIA dataset and the Cityscapes dataset is simply wider than the gap between GTA5 and Cityscapes.
This is mainly due to the fact that the viewpoint in the scenes in the SYNTHIA dataset varies a lot while Cityscapes and GTA5 mainly show scenes from the viewpoint of a pedestrian or the viewpoint of a driver. The SYNTHIA dataset also consists of less annotated images.
Common practice excludes many of the hard classes which results in $\textbf{mIoU}^{13}$ scores being very similar to the $\textbf{mIoU}^{19}$ scores that we obtain when training with the GTA5 dataset. 
The above-mentioned issues apply to most methods. However, our method outperforms other methods. Thus, in the next section, we conduct an ablation study to identify the contribution of the individual components of our method to this outperformance. 

\subsection{Ablation Study}
\label{sec:ablation}
In this section, we conduct an ablation study on various components of our proposed method. In addition, we compare different components directly to their counterparts in similar methods~\cite{choi2019self,zhou2020uncertainty}. We also analyse the remaining domain gap to get a better understanding of our results.
\paragraph{Image-to-Image Translation.}
\label{par:im2imablation}
\begin{table}
    \caption{Ablation study for the proposed image-to-image translation method from the synthetic GTA5 dataset to the Cityscapes dataset. We report mean $\textbf{mIoU}^{19}$ on the Cityscapes validation set.}
    \centering
    \begin{tabular}{l | c} 
        \hline
        Component & $\textbf{mIoU}^{19}$ \\ [0.5ex] 
        \hline
        LSGAN~\cite{mao2017least} & \textbf{43.9 } \\
        SGAN~\cite{goodfellow2014gan} & 42.9  \\
        w/o $\mathcal{L}^{D,G,clsgan}$& 42.7   \\
        w/o $\mathcal{L}^{G,seg}$ & 38.1  \\
        w/o pre-computed pseudo-labels & 40.0 \\ 
    \end{tabular}
    \label{tab:im2imablation}
\end{table}
The results of the ablation study are shown in~\autoref{tab:im2imablation}. For simplicity, we omit the \ac{SSL} part of the third stage of our method, i.e., we use the images from our \ac{I2I}~method to learn a segmentation model in a purely supervised fashion. We train on 50\% source data and 50\% translated data.
As expected, our performance is highly dependent on the semantic consistency loss $\mathcal{L}^{G,seg}$ from \autoref{eq:genloss} that we use to supervise the generator. We lose $\approx5.7\%$~$\textbf{mIoU}^{19}$ absolute performance by removing this component which demonstrates the effectiveness of our approach. Note that this ablation evaluates the impact of $\mathcal{L}^{G,seg}$. $\mathcal{L}^{D,seg}$ is still applied during training, i.e., the discriminator has access to the segmentation information but is not able to properly transfer this information to the generator. This can be improved by applying $\mathcal{L}^{G,seg}$ during training which explicitly enforces the transfer of segmentation knowledge.
Note that our method is still able to deliver $40\%$~$\textbf{mIoU}^{19}$ when using simple pseudo-labels created online during training from the output of the AC-head, i.e., if we train without pseudo-labels supplied by an external method. Nevertheless, using higher quality pseudo-labels from the proposed segmentation method increases performance by an absolute $3.8\%$~$\textbf{mIoU}^{19}$. Removing the class-wise GAN feedback reduces performance by around $1.2\%$~$\textbf{mIoU}^{19}$. Swapping the LSGAN target to a standard GAN~(SGAN) target reduces performance by $\approx 1\%$~$\textbf{mIoU}^{19}$.

\begin{figure*}
    \centering
     \begin{subfigure}[b]{0.49\textwidth}
         \includegraphics[width=\textwidth]{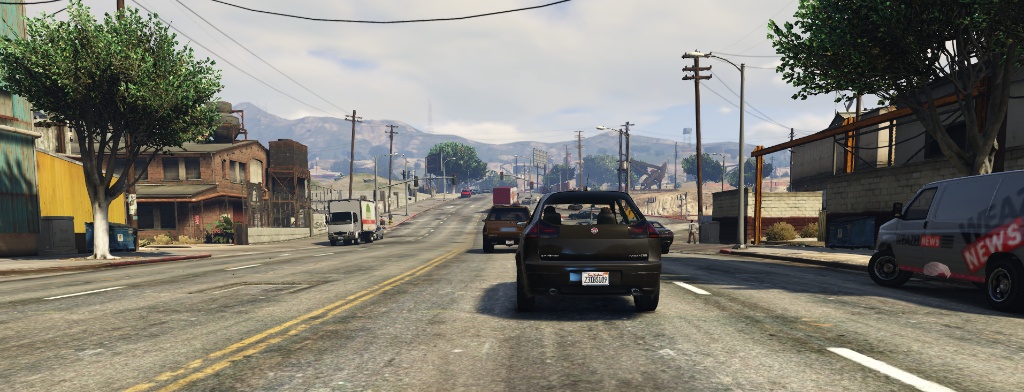}
     \end{subfigure}
     \begin{subfigure}[b]{0.49\textwidth}
         \includegraphics[width=\textwidth]{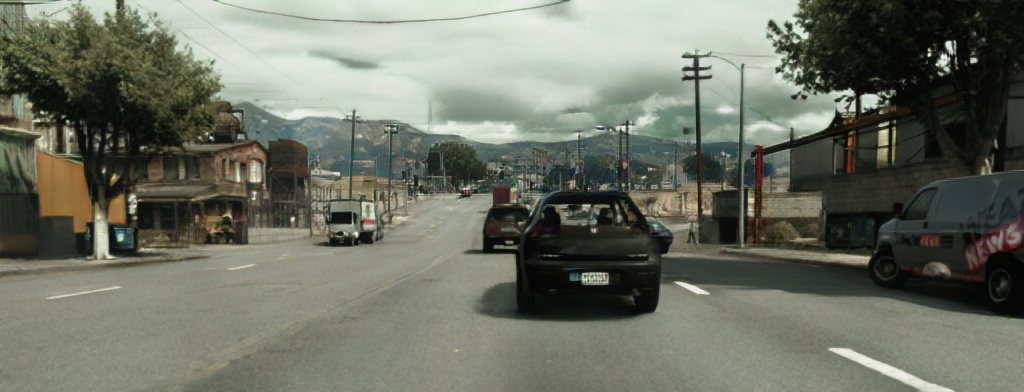}
     \end{subfigure}
    \\
     \begin{subfigure}[b]{0.49\textwidth}
         \includegraphics[width=\textwidth]{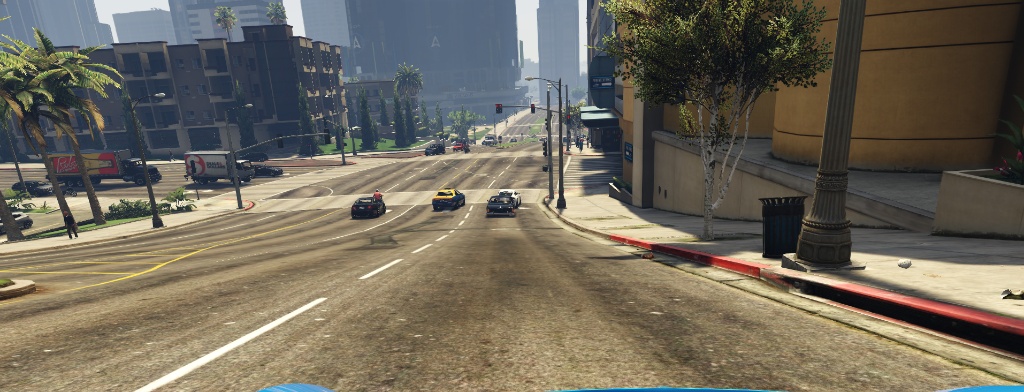}
     \end{subfigure}
     \begin{subfigure}[b]{0.49\textwidth}
         \includegraphics[width=\textwidth]{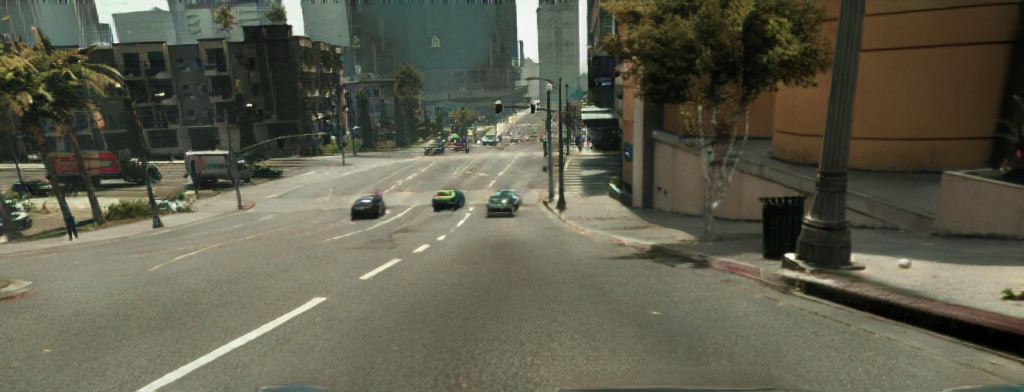}
     \end{subfigure}
      \\
     \begin{subfigure}[b]{0.49\textwidth}
         \includegraphics[width=\textwidth]{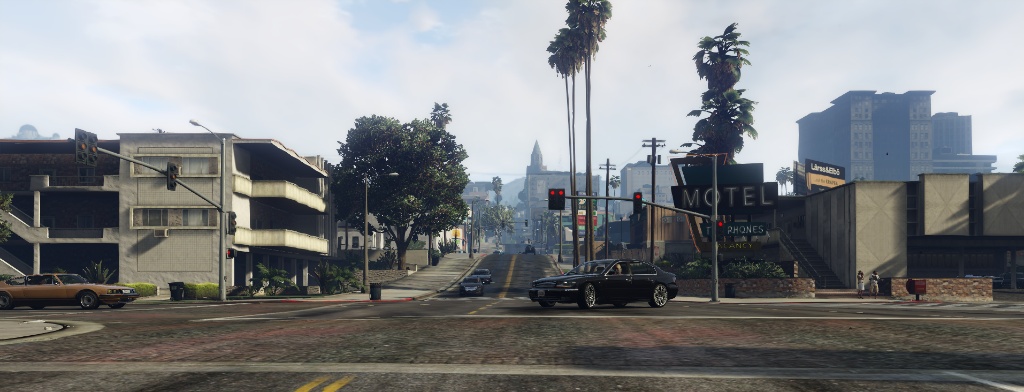}
     \end{subfigure}
     \begin{subfigure}[b]{0.49\textwidth}
         \includegraphics[width=\textwidth]{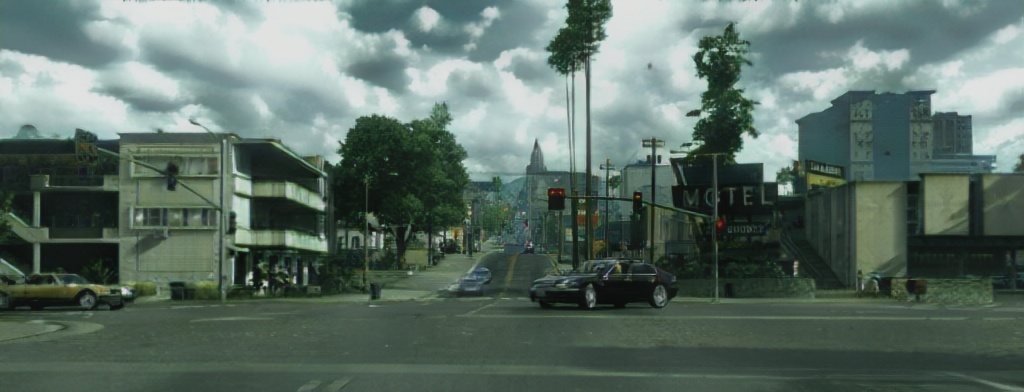}
     \end{subfigure}
     \\
    \caption{Examples of translated images. The left column shows synthetic images from the GTA5 dataset. The right column shows the corresponding translated images.}
    \label{fig:examples}
\end{figure*}

\autoref{fig:examples} shows examples of transformed synthetic images. We observe that the textures of roads and trees look much more realistic. Also, the sky is generally more cloudy which is very  characteristic for the Cityscapes dataset.

\paragraph{Comparison with Similar Methods.} 
\begin{table}
    \caption{Ablation study and comparison to existing similar work. We evaluate and compare the image translation and constancy regularization. We report results from GTA5 to Cityscapes.
    $\mathcal{L}^{F_S, seg}$ is the segmentation loss defined in \autoref{eq:Sup}. \ac{SSL} refers to the consistency loss $\mathcal{L}^{F_s,con}$ as defined in \autoref{eq:SSL}. \textit{\ac{I2I}} means that we use translated images from our proposed \ac{I2I} method.
    }
    \centering
    \begin{tabularx}{\linewidth}{l|p{3cm}@{}|c} 
        \hline
        Method & Component & mIoU  \\ [0.5ex] 
        \hline
        Source only~(ours) & $\mathcal{L}^{F_S, seg}$ & 30.5 \\ 
        \hline
        \cite{choi2019self} & $\mathcal{L}^{F_S, seg}+\ac{SSL}$ & 32.6 \\
        \cite{zhou2020uncertainty} & $\mathcal{L}^{F_S, seg}+\ac{SSL}$ & 35.6 \\
        Ours               & $\mathcal{L}^{F_S, seg}+\ac{SSL}$ & \textbf{39.2} \\ 
        \hline
        \cite{choi2019self} & $\mathcal{L}^{F_S, seg}+\textit{\ac{I2I}}$ & 35.4 \\
        \cite{zhou2020uncertainty} & $\mathcal{L}^{F_S, seg}+\textit{\ac{I2I}}$ & 35.1 \\
        Ours               & $\mathcal{L}^{F_S, seg}+\textit{\ac{I2I}}$ & \textbf{43.9} \\ 
        \hline        
        \cite{choi2019self} & $\mathcal{L}^{F_S, seg}+\ac{SSL}+\textit{\ac{I2I}}$ & 42.5 \\
        \cite{zhou2020uncertainty} & $\mathcal{L}^{F_S, seg}+\ac{SSL}+\textit{\ac{I2I}}$ & 47.8 \\
        Ours               & $\mathcal{L}^{F_S, seg}+\ac{SSL}+\textit{\ac{I2I}}$ & \textbf{50.1} \\ 
        \hline
    \end{tabularx}
    \label{tab:Comp}
\end{table}
In \autoref{tab:Comp}, we compare our two main components to the main components of similar work by Choi~\etal~\cite{choi2019self} and Zhou~\etal~\cite{zhou2020uncertainty}. We compare the components in isolation and in combination. Both components improve performance substantially upon previous work.
We can clearly see that the improvement of the \ac{I2I} method is predominantly achieved through the semantic consistency framework that we described in \autoref{sec:translation}. However, in \autoref{tab:im2imablation}, we can also clearly see, that the performance of our \ac{I2I} method is heavily impacted by the quality of the pseudo-labels. This shows that both components benefit each other.

\subsection{Domain Gap Analysis}
\begin{table}
    \caption{Domain gap evaluation. Our method closes the domain gap between GTA5 and Cityscapes by 68.3\%. }
    \centering
    \begin{tabular}{c | c| c } 
        \hline
         & $\textbf{mIoU}^{19}$ & domain gap\\
        \hline
        Cityscapes Model & 68.0 & 0.0\%  \\ 
        \hline
        Source only & 39.6 &  100.0\% \\
        Ours & 59.0 &  31.7\% \\
        \hline
    \end{tabular}
    \label{tab:i2i_upper_bound}
\end{table}
In order to estimate the remaining domain gap we retrain the linear classification layer of the segmentation model on real labels from the Cityscapes dataset. We compare to a model that is trained on Cityscapes only. The results are summarised in \autoref{tab:i2i_upper_bound}. We argue that retraining the classification layer is necessary for a proper comparison because it allows to overcome fundamental differences in class annotations between the synthetic source and the real target data. 
In essence, this means that we evaluate the quality of the learned features and their applicability to data from the target domain. In this context, feature quality refers to the linear separability of the object classes from the Cityscapes dataset in the features. This linear separability can be assessed by training a linear classifier on top of these features. Such an approach is commonly referred to as a \textit{linear probe}~\cite{bengio2017probe}.
We argue that a supervised model that is trained on annotated data from the target domain gives a reasonable upper bound on the achievable segmentation performance. The total domain gap between the target domain $\mathcal{T}$ and the source domain $\mathcal{S}$ then is the difference between the performance of this model and a model trained on source data only. Again, we retrain the linear classification layer of this source model with data from the target domain.
This source model achieves $39.6\%$~$\textbf{mIoU}^{19}$ compared to the upper bound of $68.0\%$~$\textbf{mIoU}^{19}$. Thus, we can conclude that the total domain gap between $\mathcal{S}$ and $\mathcal{T}$ is equal to $28.4\%$~$\textbf{mIoU}^{19}$ points.
In comparison, our method achieves $59.0\%$~$\textbf{mIoU}^{19}$ when retraining the classification layer. This reduces our estimate of the remaining domain gap to $\approx 9\%$. 
This equals a reduction of the domain gap by $68.3\%$ when compared to the model that is trained on source data only.

\section{Conclusion}
In this work, we have investigated the problem of unsupervised domain adaptation for semantic segmentation. We proposed two complementary approaches in order to reduce the gap between the data of a synthetic source domain and the real-world target domain. More specifically, we have shown that an adversarial image-to-image translation model that is trained with an auxiliary segmentation task on images of both domains yields significantly better results. We show that pseudo-labels can be leveraged to improve this process. The combination of the proposed methods outperforms previous state-of-the-art combinations of image-to-image translation and semi-supervised learning for domain adaptation on relevant benchmarks by a considerable margin. 
\bibliographystyle{apalike}
{
\bibliography{example}}

\begin{thebibliography}{}

\bibitem[Alain and Bengio, 2017]{bengio2017probe}
Alain, G. and Bengio, Y. (2017).
\newblock Understanding intermediate layers using linear classifier probes.
\newblock In {\em 5th International Conference on Learning Representations,
  {ICLR} 2017, Toulon, France, April 24-26, 2017, Workshop Track Proceedings}.
  OpenReview.net.

\bibitem[Chen et~al., 2018]{chen2017deeplab}
Chen, L., Papandreou, G., Kokkinos, I., Murphy, K., and Yuille, A.~L. (2018).
\newblock Deeplab: Semantic image segmentation with deep convolutional nets,
  atrous convolution, and fully connected crfs.
\newblock {\em {IEEE} Trans. Pattern Anal. Mach. Intell.}, 40(4):834--848.

\bibitem[Chen et~al., 2017]{chen2017no}
Chen, Y., Chen, W., Chen, Y., Tsai, B., Wang, Y.~F., and Sun, M. (2017).
\newblock No more discrimination: Cross city adaptation of road scene
  segmenters.
\newblock In {\em {IEEE} International Conference on Computer Vision, {ICCV}
  2017, Venice, Italy, October 22-29, 2017}, pages 2011--2020. {IEEE} Computer
  Society.

\bibitem[Chen et~al., 2019]{chen2019crdoco}
Chen, Y., Lin, Y., Yang, M., and Huang, J. (2019).
\newblock Crdoco: Pixel-level domain transfer with cross-domain consistency.
\newblock In {\em {IEEE} Conference on Computer Vision and Pattern Recognition,
  {CVPR} 2019, Long Beach, CA, USA, June 16-20, 2019}, pages 1791--1800.
  Computer Vision Foundation / {IEEE}.

\bibitem[Choi et~al., 2019]{choi2019self}
Choi, J., Kim, T., and Kim, C. (2019).
\newblock Self-ensembling with gan-based data augmentation for domain
  adaptation in semantic segmentation.
\newblock In {\em 2019 {IEEE/CVF} International Conference on Computer Vision,
  {ICCV} 2019, Seoul, Korea (South), October 27 - November 2, 2019}, pages
  6829--6839. {IEEE}.

\bibitem[Cordts et~al., 2016]{Cordts2016Cityscapes}
Cordts, M., Omran, M., Ramos, S., Rehfeld, T., Enzweiler, M., Benenson, R.,
  Franke, U., Roth, S., and Schiele, B. (2016).
\newblock The cityscapes dataset for semantic urban scene understanding.
\newblock In {\em 2016 {IEEE} Conference on Computer Vision and Pattern
  Recognition, {CVPR} 2016, Las Vegas, NV, USA, June 27-30, 2016}, pages
  3213--3223. {IEEE} Computer Society.

\bibitem[Deng et~al., 2009]{deng2009imagenet}
Deng, J., Dong, W., Socher, R., Li, L., Li, K., and Fei{-}Fei, L. (2009).
\newblock Imagenet: {A} large-scale hierarchical image database.
\newblock In {\em 2009 {IEEE} Computer Society Conference on Computer Vision
  and Pattern Recognition {(CVPR} 2009), 20-25 June 2009, Miami, Florida,
  {USA}}, pages 248--255. {IEEE} Computer Society.

\bibitem[Goodfellow et~al., 2014]{goodfellow2014gan}
Goodfellow, I.~J., Pouget{-}Abadie, J., Mirza, M., Xu, B., Warde{-}Farley, D.,
  Ozair, S., Courville, A.~C., and Bengio, Y. (2014).
\newblock Generative adversarial nets.
\newblock In Ghahramani, Z., Welling, M., Cortes, C., Lawrence, N.~D., and
  Weinberger, K.~Q., editors, {\em Advances in Neural Information Processing
  Systems 27: Annual Conference on Neural Information Processing Systems 2014,
  December 8-13 2014, Montreal, Quebec, Canada}, pages 2672--2680.

\bibitem[He et~al., 2016]{he2016deep}
He, K., Zhang, X., Ren, S., and Sun, J. (2016).
\newblock Deep residual learning for image recognition.
\newblock In {\em 2016 {IEEE} Conference on Computer Vision and Pattern
  Recognition, {CVPR} 2016, Las Vegas, NV, USA, June 27-30, 2016}, pages
  770--778. {IEEE} Computer Society.

\bibitem[Hoffman et~al., 2018]{CyCada}
Hoffman, J., Tzeng, E., Park, T., Zhu, J., Isola, P., Saenko, K., Efros, A.~A.,
  and Darrell, T. (2018).
\newblock Cycada: Cycle-consistent adversarial domain adaptation.
\newblock In Dy, J.~G. and Krause, A., editors, {\em Proceedings of the 35th
  International Conference on Machine Learning, {ICML} 2018,
  Stockholmsm{\"{a}}ssan, Stockholm, Sweden, July 10-15, 2018}, volume~80 of
  {\em Proceedings of Machine Learning Research}, pages 1994--2003. {PMLR}.

\bibitem[Hoffman et~al., 2016]{hoffman2016fcns}
Hoffman, J., Wang, D., Yu, F., and Darrell, T. (2016).
\newblock Fcns in the wild: Pixel-level adversarial and constraint-based
  adaptation.
\newblock {\em CoRR}, abs/1612.02649.

\bibitem[Hong et~al., 2018]{hong2018conditional}
Hong, W., Wang, Z., Yang, M., and Yuan, J. (2018).
\newblock Conditional generative adversarial network for structured domain
  adaptation.
\newblock In {\em 2018 {IEEE} Conference on Computer Vision and Pattern
  Recognition, {CVPR} 2018, Salt Lake City, UT, USA, June 18-22, 2018}, pages
  1335--1344. Computer Vision Foundation / {IEEE} Computer Society.

\bibitem[Karras et~al., 2018]{karras2017progressive}
Karras, T., Aila, T., Laine, S., and Lehtinen, J. (2018).
\newblock Progressive growing of gans for improved quality, stability, and
  variation.
\newblock In {\em 6th International Conference on Learning Representations,
  {ICLR} 2018, Vancouver, BC, Canada, April 30 - May 3, 2018, Conference Track
  Proceedings}. OpenReview.net.

\bibitem[Ke et~al., 2019]{ke2019dual}
Ke, Z., Wang, D., Yan, Q., Ren, J. S.~J., and Lau, R. W.~H. (2019).
\newblock Dual student: Breaking the limits of the teacher in semi-supervised
  learning.
\newblock In {\em 2019 {IEEE/CVF} International Conference on Computer Vision,
  {ICCV} 2019, Seoul, Korea (South), October 27 - November 2, 2019}, pages
  6727--6735. {IEEE}.

\bibitem[Kingma and Ba, 2015]{kingma2015adam}
Kingma, D.~P. and Ba, J. (2015).
\newblock Adam: {A} method for stochastic optimization.
\newblock In Bengio, Y. and LeCun, Y., editors, {\em 3rd International
  Conference on Learning Representations, {ICLR} 2015, San Diego, CA, USA, May
  7-9, 2015, Conference Track Proceedings}.

\bibitem[Lee et~al., 2019]{lee2019SWD}
Lee, C., Batra, T., Baig, M.~H., and Ulbricht, D. (2019).
\newblock Sliced wasserstein discrepancy for unsupervised domain adaptation.
\newblock In {\em {IEEE} Conference on Computer Vision and Pattern Recognition,
  {CVPR} 2019, Long Beach, CA, USA, June 16-20, 2019}, pages 10285--10295.
  Computer Vision Foundation / {IEEE}.

\bibitem[Lee, 2013]{Pseudo-Labels}
Lee, D.-H. (2013).
\newblock Pseudo-label: The simple and efficient semi-supervised learning
  method for deep neural networks.
\newblock In {\em Workshop on challenges in representation learning, ICML},
  volume~3, page 896.

\bibitem[Long et~al., 2015]{long2015fully}
Long, J., Shelhamer, E., and Darrell, T. (2015).
\newblock Fully convolutional networks for semantic segmentation.
\newblock In {\em {IEEE} Conference on Computer Vision and Pattern Recognition,
  {CVPR} 2015, Boston, MA, USA, June 7-12, 2015}, pages 3431--3440. {IEEE}
  Computer Society.

\bibitem[Mao et~al., 2017]{mao2017least}
Mao, X., Li, Q., Xie, H., Lau, R. Y.~K., Wang, Z., and Smolley, S.~P. (2017).
\newblock Least squares generative adversarial networks.
\newblock In {\em {IEEE} International Conference on Computer Vision, {ICCV}
  2017, Venice, Italy, October 22-29, 2017}, pages 2813--2821. {IEEE} Computer
  Society.

\bibitem[Murez et~al., 2018]{murez2018image}
Murez, Z., Kolouri, S., Kriegman, D.~J., Ramamoorthi, R., and Kim, K. (2018).
\newblock Image to image translation for domain adaptation.
\newblock In {\em 2018 {IEEE} Conference on Computer Vision and Pattern
  Recognition, {CVPR} 2018, Salt Lake City, UT, USA, June 18-22, 2018}, pages
  4500--4509. Computer Vision Foundation / {IEEE} Computer Society.

\bibitem[Odena et~al., 2017]{odena2017conditional}
Odena, A., Olah, C., and Shlens, J. (2017).
\newblock Conditional image synthesis with auxiliary classifier gans.
\newblock In Precup, D. and Teh, Y.~W., editors, {\em Proceedings of the 34th
  International Conference on Machine Learning, {ICML} 2017, Sydney, NSW,
  Australia, 6-11 August 2017}, volume~70 of {\em Proceedings of Machine
  Learning Research}, pages 2642--2651. {PMLR}.

\bibitem[Pizzati et~al., 2020]{pizzati2020domain}
Pizzati, F., de~Charette, R., Zaccaria, M., and Cerri, P. (2020).
\newblock Domain bridge for unpaired image-to-image translation and
  unsupervised domain adaptation.
\newblock In {\em {IEEE} Winter Conference on Applications of Computer Vision,
  {WACV} 2020, Snowmass Village, CO, USA, March 1-5, 2020}, pages 2979--2987.
  {IEEE}.

\bibitem[Richter et~al., 2016]{Richter2016gta5}
Richter, S.~R., Vineet, V., Roth, S., and Koltun, V. (2016).
\newblock Playing for data: Ground truth from computer games.
\newblock In Leibe, B., Matas, J., Sebe, N., and Welling, M., editors, {\em
  Computer Vision - {ECCV} 2016 - 14th European Conference, Amsterdam, The
  Netherlands, October 11-14, 2016, Proceedings, Part {II}}, volume 9906 of
  {\em Lecture Notes in Computer Science}, pages 102--118. Springer.

\bibitem[Ros et~al., 2016]{ros2016synthia}
Ros, G., Sellart, L., Materzynska, J., V{\'{a}}zquez, D., and L{\'{o}}pez,
  A.~M. (2016).
\newblock The {SYNTHIA} dataset: {A} large collection of synthetic images for
  semantic segmentation of urban scenes.
\newblock In {\em 2016 {IEEE} Conference on Computer Vision and Pattern
  Recognition, {CVPR} 2016, Las Vegas, NV, USA, June 27-30, 2016}, pages
  3234--3243. {IEEE} Computer Society.

\bibitem[Sankaranarayanan et~al., 2018]{sankaranarayanan2018learning}
Sankaranarayanan, S., Balaji, Y., Jain, A., Lim, S., and Chellappa, R. (2018).
\newblock Learning from synthetic data: Addressing domain shift for semantic
  segmentation.
\newblock In {\em 2018 {IEEE} Conference on Computer Vision and Pattern
  Recognition, {CVPR} 2018, Salt Lake City, UT, USA, June 18-22, 2018}, pages
  3752--3761. Computer Vision Foundation / {IEEE} Computer Society.

\bibitem[Simonyan and Zisserman, 2015]{simonyan2014vgg}
Simonyan, K. and Zisserman, A. (2015).
\newblock Very deep convolutional networks for large-scale image recognition.
\newblock In Bengio, Y. and LeCun, Y., editors, {\em 3rd International
  Conference on Learning Representations, {ICLR} 2015, San Diego, CA, USA, May
  7-9, 2015, Conference Track Proceedings}.

\bibitem[Taigman et~al., 2017]{taigman2016unsupervised}
Taigman, Y., Polyak, A., and Wolf, L. (2017).
\newblock Unsupervised cross-domain image generation.
\newblock In {\em 5th International Conference on Learning Representations,
  {ICLR} 2017, Toulon, France, April 24-26, 2017, Conference Track
  Proceedings}. OpenReview.net.

\bibitem[Tarvainen and Valpola, 2017]{MeanTeacher}
Tarvainen, A. and Valpola, H. (2017).
\newblock Mean teachers are better role models: Weight-averaged consistency
  targets improve semi-supervised deep learning results.
\newblock In Guyon, I., von Luxburg, U., Bengio, S., Wallach, H.~M., Fergus,
  R., Vishwanathan, S. V.~N., and Garnett, R., editors, {\em Advances in Neural
  Information Processing Systems 30: Annual Conference on Neural Information
  Processing Systems 2017, December 4-9, 2017, Long Beach, CA, {USA}}, pages
  1195--1204.

\bibitem[Toldo et~al., 2020]{toldo2020unsupervised_survey}
Toldo, M., Maracani, A., Michieli, U., and Zanuttigh, P. (2020).
\newblock Unsupervised domain adaptation in semantic segmentation: a review.
\newblock {\em CoRR}, abs/2005.10876.

\bibitem[van Engelen and Hoos, 2020]{SSL_Survey}
van Engelen, J.~E. and Hoos, H.~H. (2020).
\newblock A survey on semi-supervised learning.
\newblock {\em Mach. Learn.}, 109(2):373--440.

\bibitem[Wang et~al., 2020a]{wang2020classes}
Wang, H., Shen, T., Zhang, W., Duan, L., and Mei, T. (2020a).
\newblock Classes matter: {A} fine-grained adversarial approach to cross-domain
  semantic segmentation.
\newblock In Vedaldi, A., Bischof, H., Brox, T., and Frahm, J., editors, {\em
  Computer Vision - {ECCV} 2020 - 16th European Conference, Glasgow, UK, August
  23-28, 2020, Proceedings, Part {XIV}}, volume 12359 of {\em Lecture Notes in
  Computer Science}, pages 642--659. Springer.

\bibitem[Wang et~al., 2021]{wang2020deepHR}
Wang, J., Sun, K., Cheng, T., Jiang, B., Deng, C., Zhao, Y., Liu, D., Mu, Y.,
  Tan, M., Wang, X., Liu, W., and Xiao, B. (2021).
\newblock Deep high-resolution representation learning for visual recognition.
\newblock {\em {IEEE} Trans. Pattern Anal. Mach. Intell.}, 43(10):3349--3364.

\bibitem[Wang et~al., 2019]{wang2019symmetric}
Wang, Y., Ma, X., Chen, Z., Luo, Y., Yi, J., and Bailey, J. (2019).
\newblock Symmetric cross entropy for robust learning with noisy labels.
\newblock In {\em 2019 {IEEE/CVF} International Conference on Computer Vision,
  {ICCV} 2019, Seoul, Korea (South), October 27 - November 2, 2019}, pages
  322--330. {IEEE}.

\bibitem[Wang et~al., 2020b]{Wang_2020_CVPR_Sim}
Wang, Z., Yu, M., Wei, Y., Feris, R., Xiong, J., Hwu, W., Huang, T.~S., and
  Shi, H. (2020b).
\newblock Differential treatment for stuff and things: {A} simple unsupervised
  domain adaptation method for semantic segmentation.
\newblock In {\em 2020 {IEEE/CVF} Conference on Computer Vision and Pattern
  Recognition, {CVPR} 2020, Seattle, WA, USA, June 13-19, 2020}, pages
  12632--12641. Computer Vision Foundation / {IEEE}.

\bibitem[Wu et~al., 2018]{wu2018dcan}
Wu, Z., Han, X., Lin, Y., Uzunbas, M.~G., Goldstein, T., Lim, S., and Davis,
  L.~S. (2018).
\newblock {DCAN:} dual channel-wise alignment networks for unsupervised scene
  adaptation.
\newblock In Ferrari, V., Hebert, M., Sminchisescu, C., and Weiss, Y., editors,
  {\em Computer Vision - {ECCV} 2018 - 15th European Conference, Munich,
  Germany, September 8-14, 2018, Proceedings, Part {V}}, volume 11209 of {\em
  Lecture Notes in Computer Science}, pages 535--552. Springer.

\bibitem[Xie et~al., 2020]{Pseudo-Labels_Teacher-Student}
Xie, Q., Luong, M., Hovy, E.~H., and Le, Q.~V. (2020).
\newblock Self-training with noisy student improves imagenet classification.
\newblock In {\em 2020 {IEEE/CVF} Conference on Computer Vision and Pattern
  Recognition, {CVPR} 2020, Seattle, WA, USA, June 13-19, 2020}, pages
  10684--10695. Computer Vision Foundation / {IEEE}.

\bibitem[Yang et~al., 2020]{yang2020phase}
Yang, Y., Lao, D., Sundaramoorthi, G., and Soatto, S. (2020).
\newblock Phase consistent ecological domain adaptation.
\newblock In {\em 2020 {IEEE/CVF} Conference on Computer Vision and Pattern
  Recognition, {CVPR} 2020, Seattle, WA, USA, June 13-19, 2020}, pages
  9008--9017. Computer Vision Foundation / {IEEE}.

\bibitem[Yu and Koltun, 2016]{yu2016multi}
Yu, F. and Koltun, V. (2016).
\newblock Multi-scale context aggregation by dilated convolutions.
\newblock In Bengio, Y. and LeCun, Y., editors, {\em 4th International
  Conference on Learning Representations, {ICLR} 2016, San Juan, Puerto Rico,
  May 2-4, 2016, Conference Track Proceedings}.

\bibitem[Zhang et~al., 2017]{zhang2017curriculum}
Zhang, Y., David, P., and Gong, B. (2017).
\newblock Curriculum domain adaptation for semantic segmentation of urban
  scenes.
\newblock In {\em {IEEE} International Conference on Computer Vision, {ICCV}
  2017, Venice, Italy, October 22-29, 2017}, pages 2039--2049. {IEEE} Computer
  Society.

\bibitem[Zhou et~al., 2020]{zhou2020uncertainty}
Zhou, Q., Feng, Z., Cheng, G., Tan, X., Shi, J., and Ma, L. (2020).
\newblock Uncertainty-aware consistency regularization for cross-domain
  semantic segmentation.
\newblock {\em CoRR}, abs/2004.08878.

\bibitem[Zhu et~al., 2017]{CycleGAN}
Zhu, J., Park, T., Isola, P., and Efros, A.~A. (2017).
\newblock Unpaired image-to-image translation using cycle-consistent
  adversarial networks.
\newblock In {\em {IEEE} International Conference on Computer Vision, {ICCV}
  2017, Venice, Italy, October 22-29, 2017}, pages 2242--2251. {IEEE} Computer
  Society.

\bibitem[Zou et~al., 2018]{cbst}
Zou, Y., Yu, Z., Kumar, B. V. K.~V., and Wang, J. (2018).
\newblock Unsupervised domain adaptation for semantic segmentation via
  class-balanced self-training.
\newblock In Ferrari, V., Hebert, M., Sminchisescu, C., and Weiss, Y., editors,
  {\em Computer Vision - {ECCV} 2018 - 15th European Conference, Munich,
  Germany, September 8-14, 2018, Proceedings, Part {III}}, volume 11207 of {\em
  Lecture Notes in Computer Science}, pages 297--313. Springer.

\end{thebibliography}
\end{document}